\documentclass[10pt,conference]{IEEEtran}
\pdfoutput=1
\IEEEoverridecommandlockouts

\usepackage{blindtext}
\usepackage{calc}
\usepackage{tikz}

\usepackage[american]{babel}
\usepackage{grffile} 

\usepackage{newtxtext,newtxmath}
\usepackage{microtype}

\usepackage{amsmath}
\usepackage{amssymb}
\usepackage{bbm}
\usepackage{mathtools}
\usepackage{dsfont}
\usepackage{braket}
\usepackage{cancel}
\usepackage{slashed}
\usepackage{float}
\usepackage[scr=boondoxo]{mathalpha}
\usepackage[numbers]{natbib}
\usepackage{subcaption}

\usepackage{graphicx}

\usepackage{siunitx} 

\usepackage{ragged2e}
\usepackage{array}
\usepackage{tabularx}
\usepackage{booktabs}
\usepackage[export]{adjustbox}

\usepackage[normalem]{ulem}

\definecolor{cset-aps-blueberry}{RGB}{28,128,158}
\definecolor{cset-aps-blue}{RGB}{46,44,184}
\definecolor{cset-aps-turquoise}{RGB}{0,67,88}
\definecolor{cset-aps-limegreen}{RGB}{190,219,67}
\definecolor{cset-aps-green}{RGB}{31,138,112}
\definecolor{cset-aps-yellow}{RGB}{255,225,25}
\definecolor{cset-aps-orange}{RGB}{253,116,0}
\definecolor{cset-aps-red}{RGB}{219,0,43}

\makeatletter
\DeclareRobustCommand{\Arrow}[1][]{%
\check@mathfonts
\if\relax\detokenize{#1}\relax
\settowidth{\dimen@}{$\m@th\rightarrow$}%
\else
\setlength{\dimen@}{#1}%
\fi
\sbox\z@{\usefont{U}{lasy}{m}{n}\symbol{41}}%
\begin{picture}(\dimen@,\ht\z@)
\roundcap
\put(\dimexpr\dimen@-.7\wd\z@,0){\usebox\z@}
\put(0,\fontdimen22\textfont2){\line(1,0){\dimen@}}
\end{picture}%
}
\makeatother

\usepackage{pgfplots}
\usepgfplotslibrary{groupplots}
\pgfplotsset{compat=1.18}

\usepackage{hyperref}
\hypersetup{%
    colorlinks=true,
    linkcolor={cset-aps-red},
    linkbordercolor={cset-aps-red},
    filecolor={cset-aps-orange},
    filebordercolor={cset-aps-orange},
    citecolor={cset-aps-blue},
    citebordercolor={cset-aps-blue},
    urlcolor={cset-aps-green},
    urlbordercolor={cset-aps-green},
    menucolor={cset-aps-limegreen},
    menubordercolor={cset-aps-limegreen},
    breaklinks=true,
    pdfborderstyle={/S/U/W 2},
    pdfpagemode=UseOutlines,
    pdfstartpage={1},
}

\usepackage[noabbrev,capitalize]{cleveref}
\crefname{equation}{Eq.}{Eqs.}
\crefname{figure}{Fig.}{Figs.}
\crefname{table}{Tab.}{Tabs.}
\crefname{section}{Sec.}{Secs.}
\crefname{subsection}{Subsec.}{Subsecs.}
\crefname{subsubsection}{Subsubsec.}{Subsubsecs.}

\usepackage{lipsum}

\usepackage{placeins}
\usepackage{epsfig}
\usepackage[inline]{enumitem}
\setlist*[enumerate]{label=(\arabic*)}

\begin{document}

\title{Optimizing Quantum Circuits via ZX Diagrams using Reinforcement Learning and Graph Neural Networks
}
\author{
\IEEEauthorblockN{Alexander Mattick\IEEEauthorrefmark{1}, Maniraman Periyasamy\IEEEauthorrefmark{1}, Christian Ufrecht\IEEEauthorrefmark{1}, Abhishek Y. Dubey\IEEEauthorrefmark{1}\IEEEauthorrefmark{3},\\
 Christopher Mutschler\IEEEauthorrefmark{1}, Axel Plinge\IEEEauthorrefmark{1}, Daniel D. Scherer\IEEEauthorrefmark{1}\IEEEauthorrefmark{4}}
\IEEEauthorblockA{\IEEEauthorrefmark{1} Fraunhofer Institut für Integrierte Schaltungen IIS, Nürnberg, Germany\\
\{firstname.lastname@iis.fraunhofer.de\}
}
\IEEEauthorblockA{\IEEEauthorrefmark{3} abhishek.yogendra.dubey@iis.fraunhofer.de}
\IEEEauthorblockA{\IEEEauthorrefmark{4} daniel.scherer2@iis.fraunhofer.de}
}
\maketitle

\begin{abstract}
  Quantum computing is currently strongly limited by the impact of noise, in particular introduced by the application of two-qubit gates.
  For this reason, reducing the number of two-qubit gates is of paramount importance on noisy intermediate-scale quantum hardware.
  To advance towards more reliable quantum computing, we introduce a framework based on ZX calculus,
  graph-neural networks and reinforcement learning for quantum circuit optimization. 
  By combining reinforcement learning and tree search, our method addresses the challenge of selecting optimal sequences of ZX calculus rewrite rules.
  Instead of relying on existing heuristic rules for minimizing circuits, our method trains a novel reinforcement learning policy that directly operates on
  ZX-graphs, therefore allowing us to search through the space of all possible circuit transformations to find a circuit significantly minimizing the number of CNOT gates.
  This way we can scale beyond hard-coded rules towards discovering arbitrary optimization rules.
  We demonstrate our method's competetiveness with state-of-the-art circuit optimizers and generalization capabilities on large sets of diverse random circuits.
\end{abstract}


\section{Introduction}
\begin{figure}
    \centering
    \includegraphics[width=.9\linewidth]{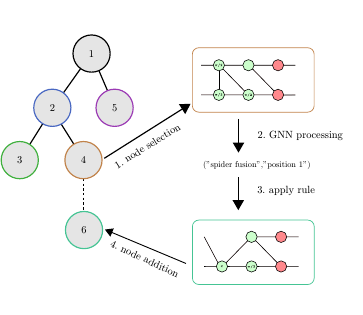}
    \caption{
    Overview of our method. Our method iteratively constructs a tree in which every node corresponds to a different transformation of the original circuit. 
    In every iteration, the agent first selects one of the transformed circuits (step 1) for further analysis.
    The chosen circuit gets processed by a GNN (step 2) which predicts both a ZX-calculus rule, and the position in the graph where the rule should be applied.
    After applying that rule (step 3) we obtain a transformed circuit. 
    The new transformation is added as a child to the selected node (step 4).
    This loop is repeated $K$ times, after which the best circuit (wrt a quality function) is selected from the tree.\label{fig:overview}
    }
\end{figure}
Quantum computing is currently limited by the strong impact of noise \cite{preskill2018quantum}. 
On superconducting hardware, errors arise mainly from two-qubit entangling gates whose execution is typically more error-prone \cite{Wagner2024} than that of single-qubit gates.
Yet, recent experimental progress has indicated that quantum computing might be at the verge of entering a utility era where first practical applications come into sight \cite{Kim2023}. 
Therefore, to accelerate the advancement towards reliable execution of quantum algorithms, it is vital to improve optimization methods that effectively decrease the two-qubit gate count.

Considerable focus has been placed on developing such optimization routines. 
Frameworks such as Qiskit \cite{contributors2023qiskit} or Tket \cite{sivarajah2020t} commonly perform optimization passes as part of the compilation process where for example neighboring gates are cancelled, commuted, or single-qubit gates are fused successively.
These frameworks also include more sophisticated techniques such as peephole optimization and template matching.
Peephole optimization \cite{osti_1785933, patel2021robust} aims to find the longest sequence of gates acting on a small subset of qubits, optimize this subcircuit, and then replace the sequence with its optimized version.
Template-matching algorithms \cite{amy2014polynomial, nam2018automated, hietala2021verified, rahman2014algorithm, iten2022exact} search for predefined subcircuits (the templates) for which simplifying circuit identities are known. 
These strategies allow optimization of quantum circuits that correspond to exponentially large matrices by application of local circuit rewrite rules.

In this work we address two main challenges present when developing optimizers based on template matching. 
Firstly, the sequence in which local rewrite rules are applied is critical for achieving optimal results. A sequence is typically determined by a selection heuristic which, however, is challenging to devise. Secondly, the templates have to be set manually, raising concerns about their generality and applicability across different scenarios.

To overcome these limitations, we introduce a framework that utilizes reinforcement learning (RL) \cite{Sutton1998} and ZX calculus \cite{Coecke2008, Coecke2010, Coecke2011}.
RL is a paradigm of machine learning where the algorithm (\textit{the agent}) learns to solve a task by interacting (\textit{carrying out actions})  with the problem (\textit{the environment}) based on a \textit{reward} signal. The agent learns a \textit{policy} that determines the action to apply given a \textit{state} of the environment. RL has proven effective in optimization problems, formulated as sequential decision tasks which require traversing local optima to reach an optimal global solution \cite{Smith2023GrowYL,Liu2022OnER,Berner2019Dota2W,Ruiz2024QuantumCO}. ZX calculus, a tensor-network representation equipped with powerful transformation rules, can represent unitary maps in a more general form than quantum circuits, potentially allowing deep optimization. Additionally, a small set of rewrite rules is sufficient to reach any equivalent diagram. 
Reinforcement learning has shown success as a selection heuristic for rewrite rules on the quantum circuit level \cite{Foesel2021}, for circuit synthesis \cite{Rietsch2024}, for node reduction of ZX diagrams \cite{Naegel2024}, and for circuit optimization \cite{Riu2024} based on a particular representation of ZX calculus known as graph-like states.  

In contrast, this work, to the best of our knowledge, is the first to make use of RL and the standard set of rules of ZX calculus for circuit optimization. 
In addition, we combine reinforcement learning with tree search, allowing the agent to reconsider previous suboptimal action selection and backtrack if necessary. 
Our method works as follows (see \cref{fig:overview}): 
We maintain a search tree over all already explored transformation sequences.
In each iteration, a quantum circuit $C_1$ is selected from the search tree and transformed into a ZX diagram, which is interpreted as a graph. 
This graph is processed by a graph neural network (GNN) and fed to the RL agent. 
Based on the current structure of the graph, at each step, the agent selects a transformation rule. 
The rule is then applied, and the new circuit is added to the tree as a child of $C_1$.
After a predefined number of steps are taken, we select the best circuit from the tree and a reward for the agent is computed (e.g.~the number of CNOT gates in the best circuit). 
During training, the reward is used to improve the agent, while during inference one can directly use the improved circuit.
Our algorithm is not limited to minimize CNOT gate count but can handle any arbitrary minimization problem,
 by replacing our CNOT-reward signal with a different one, such as, for example, a T-count minimization reward.

We demonstrate the generalization ability of our learned optimization by evaluation on a vast number of randomly generated circuits with different gate ratios (Section~\ref{sec:Experiments}).
We also showcase how our method can fit into existing optimization pipelines for large circuits using peephole optimization (Section~\ref{subsec:Peephole}).
Evaluation on random circuits ensures that our method generalizes and does not simply exploit characteristics of existing quantum circuits.
We find our algorithm to be competetive with existing manually designed rules-based optimizers, while not being constrained to any fixed rule structure.

\section{ZX-calculus}\label{sec:ZX-calculus}

ZX calculus \cite{Coecke2008, Coecke2010, Coecke2011,Wetering2020} is a tensor-network description of linear maps, and is used to represent quantum circuit. The combination with powerful local transformation rules, while keeping the underlying tensor representation of the circuit preserved, allows diagrammatic manipulation of e.g.~a quantum circuit. ZX calculus has been used for circuit optimization 
\cite{Cowtan2020, Staudacher2023, Kissinger2020a, Duncan2020, Holker2024},  error correction \cite{Chancellor2023,Garvie2018}, equivalence checking \cite{Peham2022}, and  circuit cutting \cite{Ufrecht2023}.
The basic elements of ZX calculus are Z-spiders
\begin{align}
\label{Z_spider}
{\scriptstyle m}\;\vcenter{\hbox{\includegraphics[width=0.9cm]{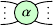}}}\;{\scriptstyle n}
\;\;&= \underbrace{|0...0\rangle}_{\scriptstyle n} \underbrace{\langle 0...0|}_{\scriptstyle m} +\mathrm{e}^{i\alpha }|1...1\rangle \langle 1...1|\\
\intertext{and X-spiders}
\label{X_spider}
{\scriptstyle m}\;\vcenter{\hbox{\includegraphics[width=0.9cm]{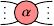}}}\;{\scriptstyle n}
\;\;&=\underbrace{|{\scriptstyle+...+}\rangle}_{\scriptstyle n}\underbrace{\langle {\scriptstyle+...+}|}_{\scriptstyle m}+\mathrm{e}^{i\alpha}|{\scriptstyle-...-}\rangle\langle {\scriptstyle-...-}|
\end{align}
where $|\pm\rangle=(|0\rangle \pm |1\rangle)/\sqrt{2}$ and $m$ and $n$ denote the number of in and outgoing wires. Typically the Hadamard gate is given an additional symbol

\begin{figure*}[t!]      
    \includegraphics[width=1.0\linewidth]{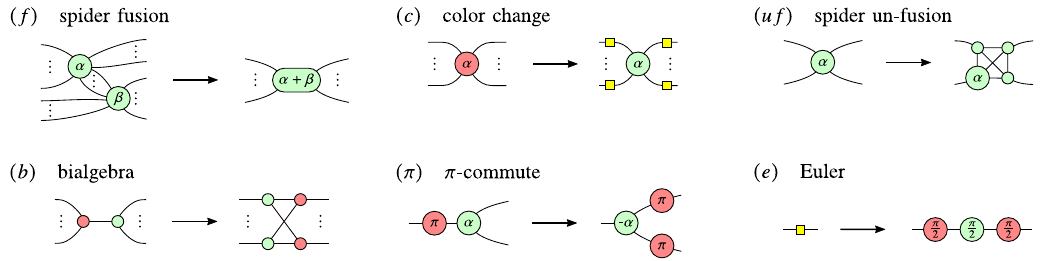}
    \caption{The figure shows the rules available to the reinforcement-learning agent in this work. The arrows show in which direction the rules can be applied. ($f$) Spider fusion: Two spiders with the same color that are connected by at least one wire can be fused and the phases are added. ($uf$) Spider un-fusion: This transformation allows to partially reverse the spider-fusion rule. For a node with $n$ in or outgoing wires with $n>3$, the rule replaces the node with a complete graph with $n$ nodes, one of them holding the initial phase. By application of a sequence of spider-fusion steps, the agent can create any layout of two nodes whose fusion would result in the left-hand-side, at least in the case of zero phase. Note that our implementation currently does not support splitting of the phase. ($\pi$) $\pi$-commute: A spider with phase $\pi$ can be pushed through a spider of different color together with a change of the sign of the spider's phase. ($c$) color change: The color of a node can be changed by pushing a Hadamard gate on each in and outgoing wire. ($b$) bialgebra: This rule can be used to interchange X and Z spiders at the cost of adding many new nodes and edges. While this rules is very powerful, it typically strongly modifies the structure of the graph, making circuit extraction challenging. ($e$) Euler rule: A Hadamard gate can be decomposed into a sequence of spiders with phase $\pi/2$ of alternating color. All rules are implemented also for X and Z spiders swapped.  In addition to those rules, we remove identity spiders (spiders with exactly two in or outgoing wires and zero phase) after each transformation step. Finally, we automatically remove any pair of wires connecting two spiders of different color. This rule can be derived from the  copy rule \cite{Wetering2020} which would have to be added together with a generalized un-fusion rule to achieve completeness of the Clifford fragment of ZX calculus. Completeness in the general sense additionally requires a modification of the Euler rule \cite{Vilmart2019}.}    
    \label{fig:rules}         
\end{figure*}

\label{Hbox}
\begin{equation}
\vcenter{\hbox{\includegraphics[width=2.54cm]{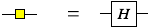}}}\,.
\end{equation}
For the sake of increased clarity, we present all relations and equalities only up to non-zero numeric factors.
ZX diagrams can be formed by composing Z and X spiders, either horizontally by connecting wires or by placing them vertically on top of each other which represents the tensor product. A CNOT gate enjoys a simple representation in terms of a Z-spider and an X-spider
\begin{equation}
\label{CX_gate}
\vcenter{\hbox{\includegraphics[width=2.56cm]{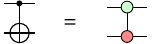}}}\,.
\end{equation}
 Being able to construct CNOT gates and single qubit rotation gates,  any quantum circuit can be represented as a ZX diagram. But even if a diagram corresponds to a unitary matrix,  it is generally a hard task to extract the corresponding quantum circuit \cite{Beaudrap2022}. Yet,  powerful extraction algorithms have been developed \cite{Backens2021} when the diagram satisfies certain graph-theoretic conditions, as explained in more detail in \cref{Circuit extraction}. 
 A set of rules is said to be complete if for each pair of diagrams that correspond to the same linear map, a sequence of rules exist which allows to transform these diagrams into each other \cite{Backens2014,Vilmart2019,Ng2017,Jeandel2018}.
The set of rules we will make use of is illustrated in \cref{fig:rules} with a detailed description of the individual rules given in the caption of the figure.
With minor modifications and additions as explained in the caption of \cref{fig:rules}, the rules are universal for the Clifford fragment of the ZX calculus \cite{Backens2014}, that is diagrams with phases that are integer multiples of $\pi/2$. 

From this discussion the potential advantage of using ZX calculus over standard template-based circuit optimizers is apparent: 
Completeness ensures the existence of a transformation sequence to any other equivalent diagram including the optimal one with respect to the optimization objective.
A complete  set of rules can be relatively small, which typically benefits the convergence of the reinforcement-learning algorithm. 
The decision on which rules to include in our model is driven by the following rationale: Our objective is to incorporate a maximum number of rules from a complete set.
At the same time the number of actions the agent can carry out at each step should be minimal which benefits the training of the agent.
Furthermore, the number of newly added nodes and edges at each transformation step should be minimal for computational cost reduction of processing the ZX diagram by the graph neural network. 
The trade-off is found by empirical performance comparisons for different subsets of rules.

\section{Reinforcement learning model}
\label{sec:rl_model}
\subsection{Formulating ZX Graph Optimization as an RL Problem}
While the ZX calculus can represent arbitrary transformations of the input circuit, using the ZX calculus to get `better' circuits (for instance lower CNOT-gate count) is highly nontrivial.
For this reason, we propose a method utilizing Reinforcement Learning (RL) to directly learn general rewriting techniques. 
Reinforcement learning has recently shown great promise in black-box decision making problems such as robotics~\cite{Smith2023GrowYL}, real time strategy games~\cite{Liu2022OnER,Berner2019Dota2W}, and also quantum circuit optimization~\cite{Ruiz2024QuantumCO, Foesel2021}.

\begin{figure*}
    \includegraphics[width=1.0\linewidth]{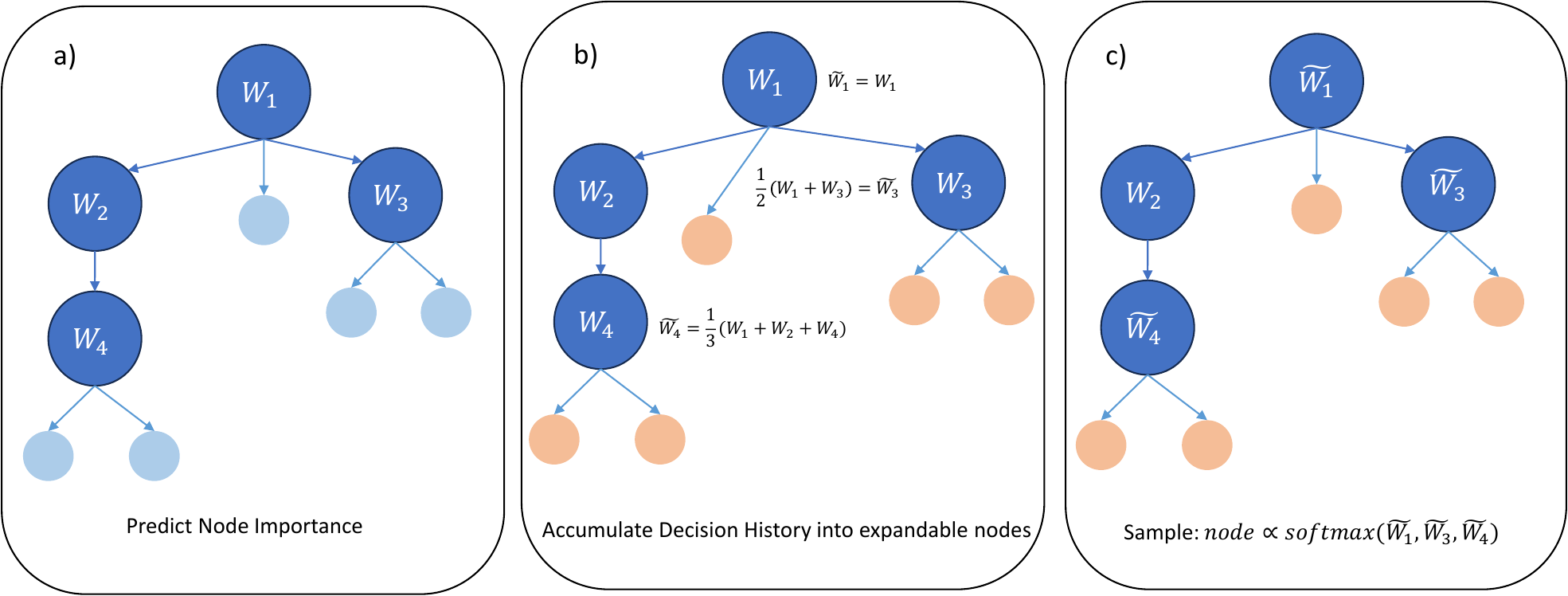}
    \caption{The node selection process. First, the model predicts a weight $W_i$ for every node in isolation (a). We then accumulate the independent weights into path-weights from the root to each node with selectable actions (b).
    We then sample the state to explore from these accumulated weights by applying a softmax operation over weights (c).
    This is mathematically equivalent to imposing a random-walk from the root to a node under a probability distribution dependent on the $W_i$.
    After a node $n$ is sampled it gets expanded according to the rule and position produced by policy $\pi_{\text{zx}}(\text{rule}|n)$}
    \label{fig:nodeExpansion}
\end{figure*}

We consider the process of optimizing a ZX graph as an RL problem acting on ZX calculus rules.
Reinforcement learning requires the problem to be formulated in the structure of a Markov Decision Process (MDP).
An MDP is characterized by a tuple $(S,\mathcal{A},T,R,p_0,\gamma)$ where $S$ is a set of states emitted from the black-box environment, $\mathcal{A}$ is a set of freely chooseable actions that can steer the sequential process, $T(s_{t+1}|s_t,a_t):S\times\mathcal{A}\times S\to[0,1]$ is the probability of transitioning from $s_t$ to $s_{t+1}$ after executing action $a\in\mathcal{A}$, $R(s_t,a_t):S\times\mathcal{A}\to\mathbb{R}$ is the immediate reward one receives when executing action $a\in\mathcal{A}$ in state $s\in S$, $\gamma$ is the discount factor weighting current against future rewards, and $p_0$ is an initial distribution over states.

Reinforcement learning aims to find a decision policy $\pi$ such that the expected discounted cumulative reward is maximized, without having any knowledge on the dynamics $T$:
\begin{equation}
    \pi^\star =\operatorname{argmax}_\pi \mathbb{E}\left[\sum^{t_\text{max}}_{t=0} \gamma^t R(s_{t+1},a_t)\pi(a_t|s_t)T(s_{t+1}|s_t,a_t)\right]
\end{equation}
For more general information regarding reinforcement learning, we recommend \cite{Sutton1998}.

Specifically, for our problem, we defined $\mathcal{A}$ as the set of rules and the positions they can be applied at, i.e., a possible action might be `apply the bialgebra rule at position ABC'.
The state $s\in S$ of the MDP is the ZX-graph, and the transition function is given by transforming the state according to the predicted action.

\subsection{Training and Inference in RL-based ZX Rewriting}
Just like in other areas of machine learning, reinforcement learning is usually applied in two phases:
First, during the training phase parameters are optimized to find the optimal policy $\pi^\star$. 
This is done by interacting with the environment and producing state-action-reward sequences (``trajectories'') and then using those trajectories to improve towards $\pi^\star$.
In the second phase (inference) the acting policy $\pi^\star$ is frozen and no further optimization takes place.
For our method, the first phase is performed on a large set of randomly generated circuits.
The second phase is the actual optimization of the target network, where the pre-trained policy can be used without further parameter updates.

Since RL considers the overall cumulative reward, rather than just the immediate improvement after applying a rule, this allows our policy to represent more complex non-monotonically improving rewriting schemes.
Due to the fact we use an in principle known to be complete calculus, we can both be sure that a graph will represent the same unitary after applying a rule, and that every possible rewriting can be represented in our RL model.
Further, since RL is capable of working with black-box models, we can include effects outside the calculus, such as circuit extraction, inside our objective function.
This simply involves setting the reward function $R$ to e.g., the number of CNOTs post extraction.
The agent will try to produce circuits with minimal CNOT-gate count after extraction by compensating for specific properties of the extraction function.
In general circuit extraction can be highly nontrivial and is discussed  in \cref{Circuit extraction}.
We implement our ZX rewriting and circuit extraction environment on top of PyZX~\cite{Kissinger2020b}.

To efficiently represent the input circuit, we use graph neural networks (GNN) \cite{Bronstein2021GeometricDL}. 
This gives a quite natural representation that is well aligned to the ZX-calculus graph.
However, instead of representing nodes and edges by their features directly (e.g.~color and phase, Hadamard/non Hadamard), we choose to represent each element by the actions admissable to that element.
For instance, if two differently colored spiders are connected, the nodes get a `can apply bialgebra rule' feature.
Representing networks this way has two advantages: Firstly, the model does not have to deduce applicable rules from features alone, secondly our model is naturally invariant to swapping the colors of all spiders.

\subsection{Tree-Based Search Strategies}
Because individual rewriting rules can increase CNOT gate count by an arbitrarily high amount, we represent the problem in the form of a tree that allows for backtracking.
Prior work on optimizing quantum circuits has relied on Monte-Carlo tree search (MCTS)~\cite{Ruiz2024QuantumCO}, which also explores a tree.
However, instead of keeping a single tree during optimization, MCTS needs to explore a tree, commit to a single action, and then build a new tree from that starting position.
This makes MCTS less efficient than our method since we simply maintain a single tree throughout optimization.
Further, since our method produces a probability over the tree, we can use modern RL solvers, such as PPO~\cite{Schulman2017ProximalPO}.

Our reinforcement learning method follows the approach of ~\cite{Mattick2023ReinforcementLF}, which maintains a search tree and explores it by producing a probability distribution over possible tree-state selections.
In this, we evaluate three learned functions - each parameterized by a graph neural network - on all graphs in the search tree.
In every step of the algorithm, we first predict a states-importance weight $W$ independently for each state in the tree (\cref{fig:nodeExpansion}a).
All state-weights are then summed from the root to the state in the tree, giving a history-aware selection priority (\cref{fig:nodeExpansion}b). 
We then sample a state to explore using a softmax over the weights (\cref{fig:nodeExpansion}c). This procedure is equivalent to randomly walking down the tree with probabilities proportional to $\exp(W)$.
Once a state is selected, we apply an action $a\in\mathcal{A}$ to that state and store the resulting graph as a child of the selected state.

For every node $n\in N$ of the $\text{tree}=(N,E)$, we predict a value function $V(n)\in\mathbb{R}$ (used in actor-critic methods, such as \cite{Schulman2017ProximalPO}), a weight function $W(n)\in\mathbb{R}$ (to implicitly induce the node-selection policy) and a policy distribution $\pi_{\text{zx}}(\text{action}|n)\in\mathbb{R}^d$ (to select the ZX-rule) for every node $n$.
The probability of selecting an individual node is proportional to the expected weight from a node to the root
\begin{equation}
    \tilde W(n) = \frac{1}{|P(r,n)|} \sum_{u \in P(r,n)} W(u)\,,\label{eq:Weight}
\end{equation}
where $P(r,n)$ is the path from root $r$ to node $n$.
Selection of the node is performed in accordance to sampling from a node-selection policy $\pi$ induced by
\begin{equation}
    \pi_{\text{selection}}(n|\operatorname{tree}) = \operatorname{softmax}\left(\{ \tilde W(n) | \forall n\in\mathscr{C}\right\})\,,\label{eq:policy}
\end{equation}
where $\mathscr{C}$ is the set of graphs in the tree that still have applicable actions.
Finally, we have an additional policy $\pi_{\text{zx}}(\text{action}|n)$ for selecting a specific rewriting rule and location in the selected node.
The overall likelihood $\pi_{\text{full}}$ is now
\begin{equation}
    \pi_{\text{full}}(\text{action}|\text{tree}) \propto \pi_{\text{zx}}(\text{action}|n)\pi_{\text{selection}}(n|\operatorname{tree})\,.
\end{equation}

In words, our policy is composed of two distributions that first sample the state in the tree to expand, and then what rule to apply for expanding the state.
Perhaps a different way of viewing this is as the model applying two actions sequentially:
First one selects the right node from a set of candidates (first action), and second one chooses what action to apply to the chosen candidate (second action).
Many RL algorithms utilise a value function $V(\cdot)$ to stabilize training, just like in \cite{Mattick2023ReinforcementLF}, we can treat $V(\text{tree})$ as a max-pooling over individual node-values:
\begin{equation}
    V(\text{tree}) = \max_{n\in N} V(n)\,.
\end{equation}
Since we have both a policy distribution $\pi_{\text{full}}$ and a value function, we can apply standard reinforcement learning algorithms, such as PPO~\cite{Schulman2017ProximalPO} to solve this MDP.
Our implementation is based on a modified version of CleanRL~\cite{Huang2021CleanRLHS}.

Since we work on the tree-MDP, we can simply select the highest-reward circuit from the tree, while keeping the Markov Property intact.
Specifically, the reward is given by 
\begin{equation}
    R(\text{tree}) = \max\left\{1-\frac{\operatorname{CNOT}(n)}{\operatorname{CNOT}(r)}\; \vline\; \forall n\in N, \text{tree}=(N,E)\right\}\,,
\end{equation}
where $\operatorname{CNOT}(n)$ is the number of CNOTs in the circuit contained in node $n$, and $r$ is the root of the search graph.
This reward amounts to minimizing the number of CNOTs in the best circuit of the current tree, normalized by the CNOT-gate count of the unmodified circuit (stored in the root).

We train the weight function $W$ and value function $V$ on a set of randomly generated circuits.
Once the training phase is over, both $W$ and $V$ are frozen and can be used to predict the optimal tree-search policy without needing any further training.

\section{Experiments}\label{sec:Experiments}

\subsection{Experimental Setup}

\begin{figure*}[htbp!]
    
    \centering
    \begin{tikzpicture}
        \begin{groupplot}[
            group style={
                group size=4 by 2,
                horizontal sep=0.5cm,
                vertical sep=1.25cm,
                xticklabels at=edge bottom,
                yticklabels at=edge left
            },
            width=0.3\linewidth,
            ybar,
            xtick distance=1
            ]
            
            \nextgroupplot[title={Level 1}, ylabel={Bin counts}, 
            xtick={0, 20, 40, 60, 80}, ymax=450]
            \addplot+[
                hist={
                    bins=80,
                    data min=0,
                    data max=80,
                }
            ] 
            table[y=level1 - no swap, col sep=comma] {figures/common_dataset/results.csv};
            
            \nextgroupplot[title={Level 2}, 
            xtick={0, 20, 40, 60, 80}, ymax=450]
            \addplot+[
                hist={
                    bins=80,
                    data min=0,
                    data max=80,
                }
            ] table[y=level2 - no swap, col sep=comma] {figures/common_dataset/results.csv};
            
            \nextgroupplot[title={Level 3}, 
            xtick={0, 20, 40, 60, 80}, ymax=450]
            \addplot+[
                hist={
                    bins=80,
                    data min=0,
                    data max=80,
                }
            ] table[y=level3 - no swap, col sep=comma] {figures/common_dataset/results.csv};
            
            \nextgroupplot[title={Level 4}, 
            xtick={0, 20, 40, 60, 80}, ymax=450]
            \addplot+[
                hist={
                    bins=80,
                    data min=0,
                    data max=80,
                }
            ] table[y=level4 - no swap, col sep=comma] {figures/common_dataset/results.csv};
            
           \nextgroupplot[title={PyZX full\_reduce}, ylabel={Bin counts}, xtick={0, 20, 40, 60, 80}, ymax=450]
            \addplot+[
                hist={
                    bins=80,
                    data min=0,
                    data max=80,
                }
            ] table[y=f_r - count SWAPS as CX, col sep=comma] {figures/common_dataset/results.csv};
            
            \nextgroupplot[title={Staudacher et al.}, 
            xtick={0, 20, 40, 60, 80}, ymax=450]
            \addplot+[
                hist={
                    bins=80,
                    data min=0,
                    data max=80,
                }
            ] table[y=korbinian, col sep=comma] {figures/common_dataset/results.csv};
            
            \nextgroupplot[title={RL agent}, 
            xtick={0, 20, 40, 60, 80}, ymax=450]
            \addplot+[
                hist={
                    bins=80,
                    data min=0,
                    data max=80,
                }
            ] table[y=agent optimized, col sep=comma] {figures/common_dataset/results.csv};
            
            \nextgroupplot[title={Brute forcing}, 
            xtick={0, 20, 40, 60, 80}, ymax=450]
            \addplot+[
                hist={
                    bins=80,
                    data min=0,
                    data max=80,
                }
            ] table[y=brute force optimum, col sep=comma] {figures/common_dataset/results.csv};
            
        \end{groupplot}
    \end{tikzpicture}
    \caption{Histogram of two-qubit gates in circuits optimized by different methods. All circuits are from the dataset with a gate ratio of 1.0/0.0/0.0/0.0 consisting of 80 CNOT gates. The bins here show the number of two-qubit gates after optimization.}
    \label{img:results_pureCnot}
\end{figure*}
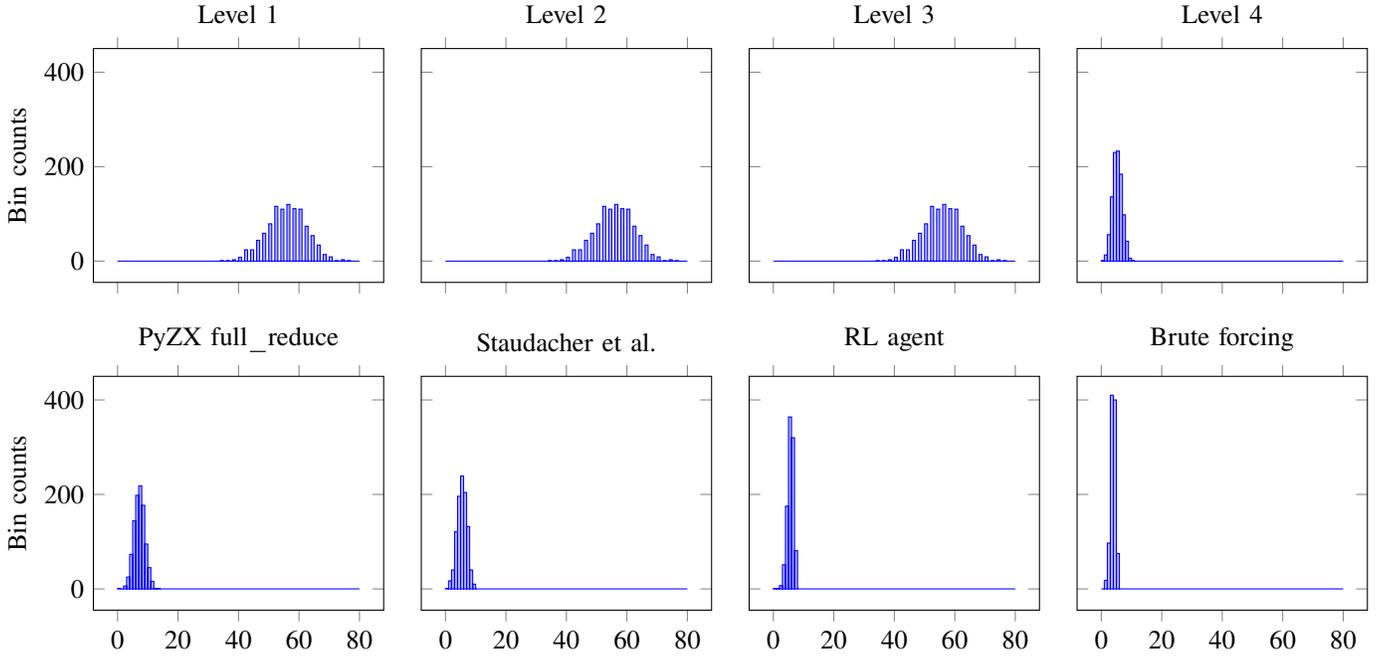

Our quantum circuit optimization environment leverages the graph structure of the ZX diagram of the given quantum circuit.
To do this we jointly train two GNN models, one for extracting the priority of the graph inside the search tree, and a second one for selecting the rule and position inside the selected ZX diagram.
We use the Pytorch Geometric API \cite{Fey_2019} to extract features from the given ZX graph representation of the quantum circuit. We use two GNN models, each consisting of 4 GCN layers~\cite{Kipf2016SemiSupervisedCW} with width $16\times\text{number of rules}$.
The first model is used to predict both the rule and the position inside the ZX graph where the rule should be applied to (see also \cref{ap:GNN}).
The second GNN parametrizes a tree-level policy by treating the ZX graph in its entirety as a single node, and propagating information through the search tree consisting of multiple equivalent ZX graphs (see~\cref{fig:overview}). To do this, the tree-level policy aggregates information of the individual graphs into two scalars per graph using mean pooling.
The first of those scalars is the value, the second the weight, both of which are needed to compute the PPO update (see \cref{sec:rl_model}).
The reinforcement learning agent was trained to reduce the two-qubit gate count in randomly generated five-qubit quantum circuits consisting of CNOT, H, $R_x$, and $R_z$ gates. To understand the learning behavior of the agent, its performance when trained with different ratios of single and two-qubit quantum gates was studied. A detailed overview of the ratios used is given below. The agent was trained for one million training steps with eight environments running in parallel. We use a batch size of 128, learning rate $3\cdot 10^{-4}$, and $\gamma = 0.99$ during the course of the training. Furthermore, to push the agent towards making hard choices, we use entropy regularization with a coefficient of $10^{-5}$. Our agent is given a tree-exploration budget of $128$ steps.
In general, when applicable, we try to stay close to the guidance in \cite{shengyi2022the37implementation} when tuning our implementation.

\subsection{Circuit Extraction}

The RL agent optimizes the ZX diagram of the given quantum circuit. The end goal of the optimization in this study is the reduction of two-qubit gates in the given quantum circuit. Yet, converting the ZX diagram to a quantum circuit representation is not trivial. In order to ensure circuit extraction, post-processing the agent-optimized ZX diagram might become necessary. However, strong post-processing will modify the ZX diagram to a greater extent, resulting in a further increase or decrease in the number of two-qubit gates. In order to get away with as little post-processing as possible, we divided the degree of post-processing into four levels. Each level increase corresponds to a stronger post-processing resulting in a greater modification to the ZX diagram. The circuit extractor applies higher levels of post-processing if and only if the current level fails to extract a valid quantum circuit from the ZX diagram. The task of the agent is to perform optimizations to the ZX diagram in such a way that the extracted circuit stays as close as possible to the optimized ZX diagram i.e., lower-level optimizations are used by the circuit extractor to extract the circuit. Even in scenarios where lower-level extractions are not viable, the optimization from the agent should lead to a reduction in two-qubit gates after extraction using higher levels. A more detailed explanation of the workings of this circuit extractor and its corresponding levels are given in \cref{Circuit extraction}.

\subsection{Training and Validation dataset}
\label{sec:training_data}

Random circuits with different CNOT, H, $R_x$, and $R_z$ rotation gates were used to sample quantum circuits, which were then converted to ZX graphs, which were then used to train the agent for optimizing the circuit. 
Random five qubit circuits consisting of CNOT/H/$R_x$/$R_z$ in the ratio of 0.6/0.2/0.1/0.1 (hereinafter referred to as dataset (i)) were used to train the optimizer for better generalization capabilities. To demonstrate the generalization capabilities of our method, we also report different gate ratios for CNOT/H/$R_x$/$R_z$: 1.0/0.0/0.0/0.0 on four qubit circuits (hereinafter referred to as dataset (ii)), and 0.25/0.25/0.25/0.25 on five qubit circuits (hereinafter referred to as dataset (iii)). The ratio of (ii) leads to a circuit consisting of two-qubit gates alone. A circuit made of just two-qubit gates is the most basic setup for two-qubit gate optimization. Here one can compare the performance of the RL agent against the optimum obtained by brute forcing (assuming a small number of qubits) following Nash et al. \cite{Nash_2020}. As the two-qubit gate ratio reduces in the input circuit, the possibilities for optimization reduce as well, making the optimization problem harder. The RL agent was trained using the dataset (i) and validated using the other datasets.

\subsection{Results}


This section presents the results of all the experiments conducted in this study. We trained a single agent to optimize the ZX graphs using the dataset (i) described in \cref{sec:training_data}. The performance of the agent was compared against the `PyZX full\_reduce' optimizer and the circuit extractors introduced in \cref{Circuit extraction}. 
The comparison against different levels of the circuit extractor is considered because the circuit extraction is done via the transformation of the ZX diagram to graph-like states (see \cref{Circuit extraction}), where nodes are fused wherever possible. 
Consequently, even without optimization prior to extraction we already find CNOT gate reduction due to non-trivial gate cancellations.
Hence the circuit extractor at different levels acts as the baseline performance metric that can be attained via simple non-trivial heuristic cancellations. 
Further, we also compare the performance of the agent against the optimization capabilities of the heuristics proposed by Staudacher et al. \cite{Staudacher2023}.

\cref{tab:results_0.6Cnot} shows the two-qubit optimization capabilities of the agent, the full\_reduce method from PyZX, and the heuristics proposed by Staudacher et al. along with the circuit extractor with various levels as baseline metric on the training dataset, dataset (i). All the results of the RL agent in this work involve the agent optimizing the ZX diagram and the circuit being extracted at the minimal circuit extractor level. Comparing the results against simple extraction by the circuit extractor shows that the optimization by the RL agent enhances the performance of the circuit extractor in terms of two-qubit gate count reduction. \cref{tab:results_0.6Cnot} shows that the agent exhibited the best performance compared to other optimizers and baseline metrics from the circuit extractor. The baseline metrics presented show that the reduction in two-qubit gates is rooted in the optimization by the agent and not primarily via the circuit extractor, as the circuit extraction after agent optimization resulted in 5 to 12 additional CNOT gate reductions compared to baseline metrics. The average extraction level used by the circuit extractor to extract the graphs optimized by the agent is 3.539 ± 0.74, and the agent outperforms all the baseline metrics. This shows that during the course of training, the agent also learns the capabilities of the circuit extractor and optimizes the circuit in such a way that the optimized graph will lead to circuit extraction with the best-suited level.

\begin{table}[htbp!]
    \centering
\caption{Two-qubit optimization results on circuits made of CNOT, H, $R_x$, and $R_z$ in the ratio of 0.6/0.2/0.1/01, respectively. Each circuit contains 80 gates in total, with 47-50 two-qubit gates among them.}
\begin{tabular}{cc}\toprule
Method                      & Resultant two-qubit gate count  \\ \midrule

PyZX full\_reduce           & 31.8 ± 5.7            \\
Staudacher et al.           & 27.6 ± 4.3             \\
RL agent                    & \textbf{27.3 ± 4.7}     \\      \midrule

Baseline \\ \cmidrule(lr){1-1}        

Circuit extractor (level 1) & 39.7 ± 4.9            \\
Circuit extractor (level 2) & 39.5 ± 4.9            \\
Circuit extractor (level 3) & 37.3 ± 6.9            \\
Circuit extractor (level 4) & 32.6 ± 6.1            \\\bottomrule

\end{tabular}
\label{tab:results_0.6Cnot}
\end{table}

Next, to analyze the generalization capabilities of the agent, we validated the performance of the agent on dataset (ii) and dataset (iii), which the agent has never seen during its training phase.  
 \cref{tab:results_pureCnot} and \cref{img:results_pureCnot} show the two-qubit optimization capabilities of the agent, different optimizers, baseline metrics from the circuit extractor, and the optimum obtained by brute forcing following the circuit representation used by Nash et al.~\cite{Nash_2020} on the dataset (ii). 
 \cref{tab:results_pureCnot}  shows that the agent exhibited the best performance compared to other methods except brute forcing. However, the circuit extractor used by all the methods introduces SWAP operations, and the brute force method does not. 
 These SWAP operations, which introduce three additional two-qubit gates per SWAP, can be removed by simply permuting the qubit order. 
 When the SWAP operations are removed, the  `PyZX full\_reduce' optimizer resulted in circuits with an average of 4.8 ± 1.5 two-qubit gates, the heuristic by Staudacher et al. with 4.9 ± 1.6 two-qubit gates and the RL agent with 3.7 ± 1.0  two-qubit gates. 
 The RL agent was able to exhibit a performance close to the optimum. All the results in this work are presented without removing the SWAPs as other optimizers include SWAPs by default. 
 The histograms presented in \cref{img:results_pureCnot} show the resultant two-qubit counts after optimization by RL agent, different optimizers, brute forcing, and circuit extractor with different levels as baseline metrics. 
 The more left-skewed the histogram is, the better the optimization capabilities. 

 From \cref{img:results_pureCnot}, one can see that the RL agent resulted in a distribution closer to the optimum by the brute force method compared to other methods. 
 The average extraction level used by the circuit extractor to extract the graphs optimized by the agent is 4.0 ± 0.000. 
 The circuit extraction with level four is an obvious choice here as the maximum reduction of CNOT gate count can be attained via maximum optimization and node cancellations in pure CNOT circuits. 
 The agent has likely learned this behavior and optimized the ZX diagrams accordingly.
 This also explains why our method works comparatively better on circuits in \cref{tab:results_pureCnot} than on the distribution it was actually trained on:
 Our method has learned to exploit the interplay between extraction levels and circuit makeup to reduce the number of CNOTs in pure-CNOT circuits.

\begin{table}[htbp!]
\caption{Two-qubit optimization results on circuits made of 100 \% two-qubit gates. Each circuit consists of 80 two-qubit gates. All the results presented are the mean and the standard deviation of the two-qubit gate count over the dataset after optimization.}
\centering
\begin{tabular}{cc}\toprule
Method                      & Resultant two-qubit gate count  \\ \midrule

PyZX full\_reduce           & 6.7 ± 1.7             \\
Staudacher et al.           & 6.9 ± 1.8             \\
RL agent                    & \textbf{5.1 ± 0.9}    \\
Brute force method          & \textbf{3.4 ± 0.8}     \\      \midrule

Baseline \\ \cmidrule(lr){1-1}

Circuit extractor (level 1) & 55.6 ± 6.3            \\
Circuit extractor (level 2) & 55.6 ± 6.3            \\
Circuit extractor (level 3) & 55.6 ± 6.3            \\
Circuit extractor (level 4) & 6.7 ± 1.7             \\\bottomrule

\end{tabular}
\label{tab:results_pureCnot}       
\end{table}



Finally, \cref{tab:results_0.25Cnot} presents the results of the agent trained and validated on the dataset (iii) which consists of random circuits comprised of an equal number of CNOT, H, $R_x$, and $R_z$ rotation gates. The agent again outperformed all the other optimizers and the baseline metrics in the setup. The average extraction level used by the circuit extractor to extract the graphs optimized by the agent is 1.263 ± 0.80.

\begin{table}[htbp!]
    \centering
\caption{Two-qubit optimization results on circuits made of CNOT, H, $R_x$ and $R_z$ in the ratio of 0.25/0.25/0.25/0.25 respectively. Each circuit contains 80 gates in total, with 19-21 two-qubit gates among them.}
\begin{tabular}{cc} \toprule
Method                      & Resultant two-qubit gate count  \\ \midrule

PyZX full\_reduce           & 27.0 ± 6.6             \\
Staudacher et al.           & 17.8 ± 3.8                \\
RL agent                    & \textbf{17.5 ± 3.7}       \\      \midrule

Baseline \\ \cmidrule(lr){1-1}
Circuit extractor (level 1) & 18.0 ± 3.8            \\
Circuit extractor (level 2) & 17.9 ± 3.8            \\
Circuit extractor (level 3) & 24.2 ± 6.5            \\
Circuit extractor (level 4) & 26.5 ± 6.7            \\\bottomrule
\end{tabular}
\label{tab:results_0.25Cnot}
\end{table}

\subsection{Peephole Optimization on large Circuits}\label{subsec:Peephole}

\begin{figure*}[!htb]
    \begin{subfigure}[b]{0.33\textwidth}
        \includegraphics[width=\textwidth]{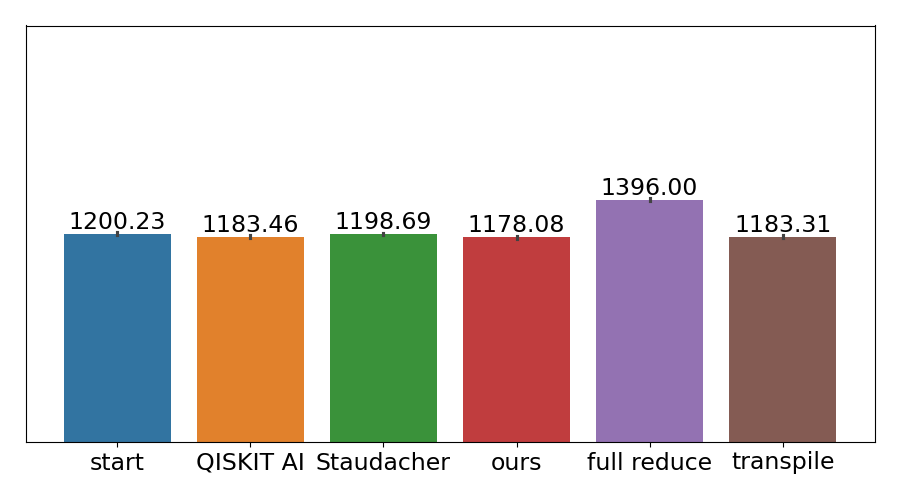}
        \caption{Random Circuits with 60/20/20 ratios}\label{fig:Peephole60}
    \end{subfigure}%
    \begin{subfigure}[b]{0.33\textwidth}
        \includegraphics[width=\textwidth]{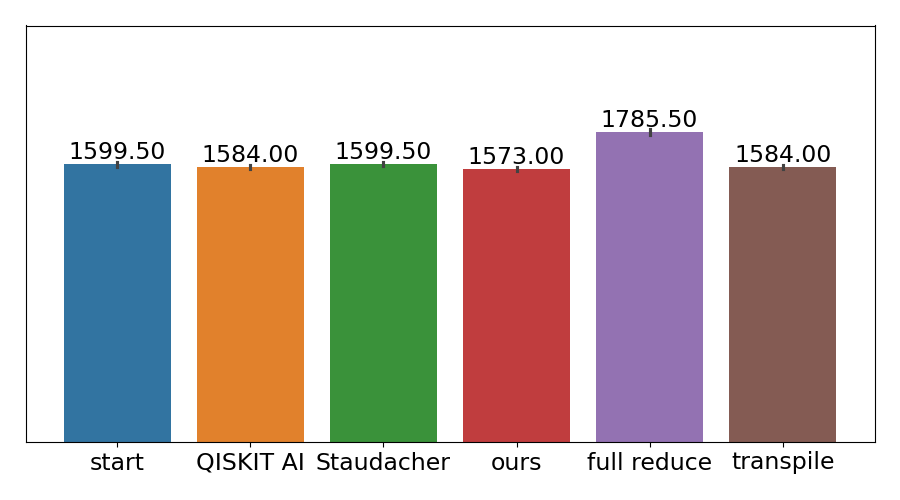}
        \caption{Random Circuits with 80/10/10 ratios}\label{fig:Peephole80}
    \end{subfigure}
    \begin{subfigure}[b]{0.33\textwidth}
        \includegraphics[width=\textwidth]{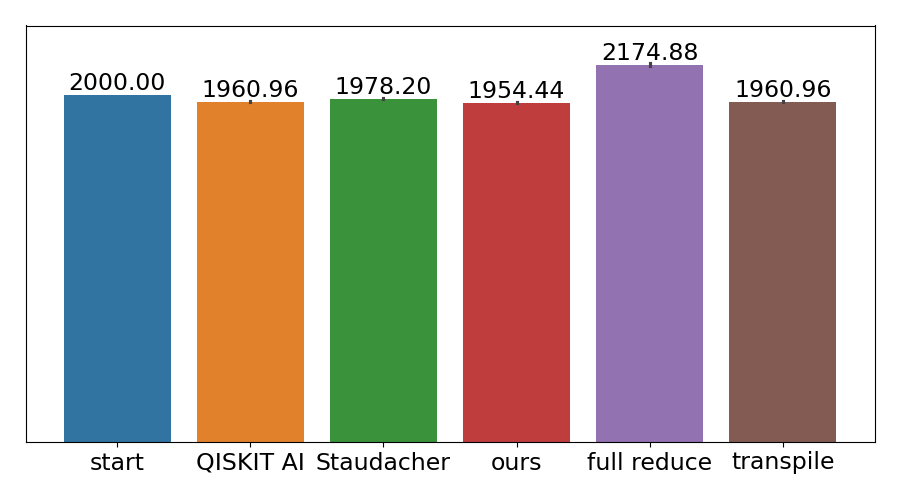}
        \caption{Random Circuits with 100/0/0 ratios}\label{fig:Peephole100}
    \end{subfigure}
    \caption{Results on fully random circuits. Both our method and \citep{Staudacher2023} were evaluated using peepholes, 
    while the qiskit transpilation and pyzx full\_reduce were evaluated on the full circuit. For qiskit transpilation we chose optimization level 3, the highest available.}
    \label{fig:PeepholeRandom}
\end{figure*}

\begin{figure*}[!htb]
    \begin{subfigure}[b]{0.33\textwidth}
        \includegraphics[width=\textwidth]{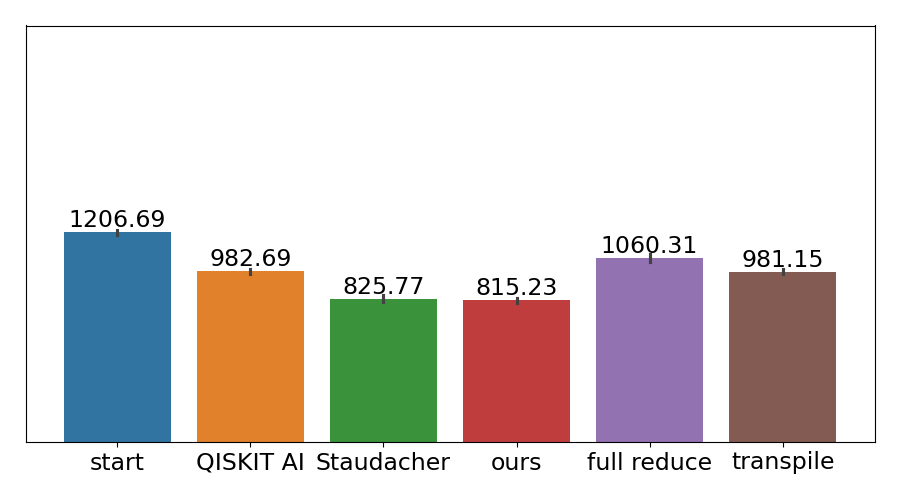}
        \caption{Assembled Circuits with 60/20/20 ratios}\label{fig:assembledPeephole60}
    \end{subfigure}%
    \begin{subfigure}[b]{0.33\textwidth}
        \includegraphics[width=\textwidth]{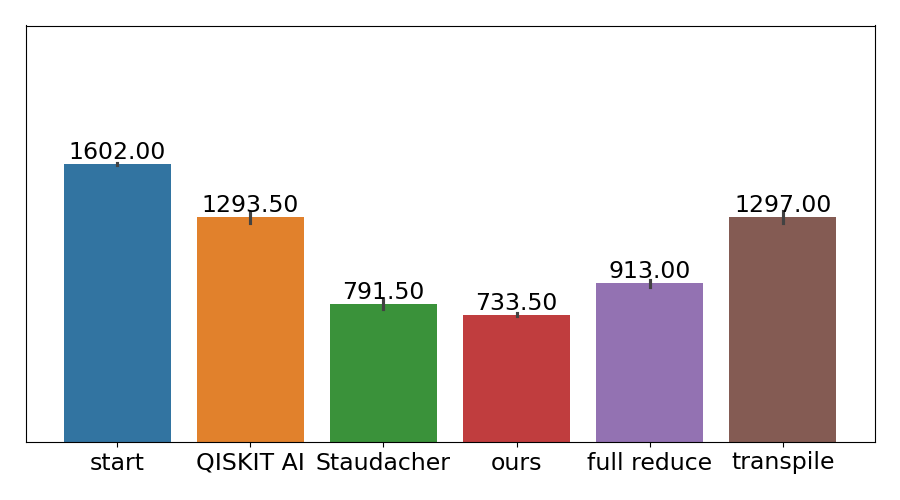}
        \caption{Assembled Circuits with 80/10/10 ratios}\label{fig:assembledPeephole80}
    \end{subfigure}
    \begin{subfigure}[b]{0.33\textwidth}
        \includegraphics[width=\textwidth]{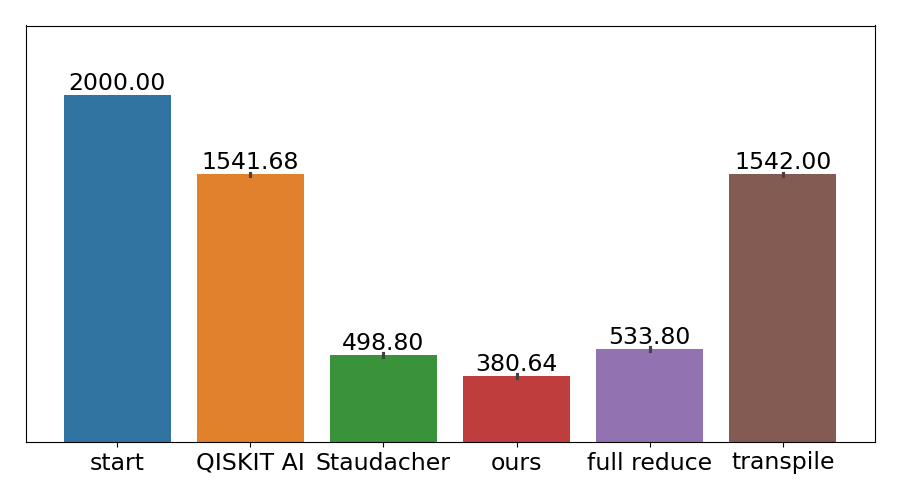}
        \caption{Assembled Circuits with 100/0/0 ratios}\label{fig:assembledPeephole100}
    \end{subfigure}
    \caption{Results on assembled random circuits. The circuits were generated by combining random 5 qubit, 50 gate subcircuits until a 50 qubit 2000 gate circuit was reached.
    Both our method and \citep{Staudacher2023} were evaluated using peepholes, 
    while the qiskit transpilation and pyzx full\_reduce were evaluated on the full circuit. For qiskit transpilation we chose optimization level 3, the highest available.}
    \label{fig:PeepholeAssembled}
\end{figure*}

To showcase the scalability of our approach we evaluate our method on large 50 qubit, 2000 gate circuits.
Since it is infeasible to train our method natively at this scale, we resort to peephole optimization and compare against
Qiskit's transpile optimization~\citep{contributors2023qiskit}, pyzx full-reduce~\citep{Kissinger2020b} and \citep{Staudacher2023}.
For partitioning we use BQskit's QuickPartitioner~\citep{BQSKit} with a target partition size of 5 qubits (same as in training).
To further increase the scalability of our method, we remove the GNN and action/position selection and replace it with a simple Multi-Layer Perceptron (MLP).
Said MLP relies on high-level features (see \cref{sec:Features}) rather than the ZX-graph, which allows the model to be significantly smaller and faster to execute.
Since we can no longer dynamically scale the action space with the GNN, we instead choose position and rule randomly amongst all legal operations and only choose which
node to continue exploring in the search tree.
Notice that this parametrization still allows the model to learn the same transformations as before, since the model simply has to re-select the same node in the search
tree until the desired action is applied.
This does lead to a worst-case increase of ``number of actions'' applications of the function, but this does not seem to lead to a measurable disadvantage in practice 
when accounting for the wall-clock time improvements gained by removing the GNN\footnote{Since removing the action selection and only keeping the tree search leads to equal performance
one might ponder whether just doing action selection and no tree search works equally well: 
We found that just training action prediction leads to a risk-averse model that only predicts ``NO-OP'' actions since such a model cannot recover from a bad rule application.}.
We presume this is the case since many optimizations are orthogonal with each other: Applying rule A to gates XYZ can be done independently from applying rule B to gates EFG,
as long as XYZ and EFG are disjoint sets.
To account for the lost performance due to retries, we explore a tree for 128 steps and restart the search 3 times with the best found solution in the current tree.
Restarting the search at the currently best node has the effect of ``pruning'' old nodes from the search that do not lead to an improvement.
Even accounting for this significantly increased search budget, the resulting method trains and infers faster than the GNN based model.

One issue with optimizing fully random circuits is the lack of structure within them, which means that the likelihood of having 
optimizable substructures is generally low:
The chance of two subsequent spiders sharing a single qubit drops linearly with the circuit width, the chance of sharing two quadratically, etc
This can be seen quite clearly in \cref{fig:PeepholeRandom}, where even in the 100\% CNOT case the originally close to optimal pyzx full\_reduce (see \cref{tab:results_pureCnot}),
and the state-of-the-art method by \citep{Staudacher2023} does not yield an improvement.
We argue this is due to the unrealistic assumption of random circuits only containing ``global'' structure, but no ``local'' structure.
Real world circuits should have both global and local structure, since large 50 qubit circuits are presumably composed of more elementary operations, 
which may only need a small subset of qubits.
We test this assumption by generating circuits that have a local structure: Instead of assembling the entire 50 qubit circuits uniformly at random,
we first generate small circuits with a width of 5 qubits and 50 gates.
These smaller circuits get assembled into large 50 qubit circuits by appending them onto a randomly chosen contiguous subset of 5 qubits, until we obtain a 50 qubit 2000 gate circuit.
By definition, these circuits now contain 5-qubit sub-operations, which should lead to more realistic quantum circuits.
Just like before, we run BQSKit's QuickPartitioner~\citep{BQSKit}, which tends to extract larger peepholes with more optimization potential, 
since we have a minimal concentration of CNOTs in every partition.
We found that this procedure often extracts larger than 50 gate peepholes since QuickPartitioner is good at merging multiple subcircuits into larger circuits.

Studying those assembled circuits in \cref{fig:PeepholeAssembled}, we observe a significant uplift in optimization performance for all methods, 
much more in line with what is expected from our results in our 5 qubit experiments.
This suggests a fundamental distribution shift between small and large random circuits, which may be interesting for future reinforcement learning training and 
Optimization benchmarking.
We find \cite{Staudacher2023} and our method to be the strongest CNOT optimizers compared to both pyzx and qiskit's transpile and AI compilation routines\footnote{We found that AI compilation does not perform well for fully connected backends.}.
It is worth mentioning though that this is not a completely unbiased comparison as pyzx's full\_reduce is not a CNOT gate optimizer and qiskit's AI compiler
is designed to also optimize for topology constrained hardware (which is not currently captured by our method).\footnote{For these benchmarks we utilize a synthetic backend with full connectivity to maintain comparability between methods}
We do still highlight these two methods due to their prominance in ZX diagram and Quantum Circuit optimization respectively.
While our method performs best in both the random and assembled random circuits, we have to note that it is also the slowest optimization algorithm.
We believe this could be changed by a more efficient implementation, as well as standard neural-network inference optimizations.
However, optimizing inference speed for neural tree-search schemes, such as ours, remains an open problem.

The experimental results demonstrated the effectiveness of using an RL agent to reduce the two-qubit gate count in a quantum circuit using ZX diagrams, GNNs, and tree search techniques. The agent outperformed other methods in the given scenarios. These results validate the robustness of our approach, highlighting its potential to enhance quantum circuit optimization. These initial results are promising, yet they represent only the first step in our research, where we study the effectiveness of our method for small quantum circuits and a fixed number of gates. There remains substantial work to be done to fully realize and expand upon the potential of our circuit optimization technique.

\section{Future work}\label{sec:FutureWork}
The results presented in this work represent the first step in exploring the potential of using reinforcement learning and the standard rule set of ZX calculus to optimize quantum circuits through GNNs and tree search. 
As an initial study, the agent did not have access to the complete set of transformations on the ZX diagrams.
However, this is to be extended so that the agent can access the complete set of ZX transformation rules to find a better or near optimal solution. 
All the results presented here were tested on four or five-qubit circuits with eighty gates per circuit. 
Graph neural networks are constructed such that they are not constrained to a predefined graph size and are known for their great generalization capabilities. 
We expect these generalization capabilities to allow training on a small graph but optimizing circuits of much larger size with the trained agent. 
This will be the focus of future work. 
Currently, the agent is trained to reduce the two-qubit gates by looking at the unitary of the full circuit as ZX graphs; 
however, this can be adapted towards other optimization goals such as T-count minimization and routing. 
A completely orthogonal area for further investigation is the design of realistically structured random circuits:
As we observed in \cref{subsec:Peephole}, simply sampling large random circuits will lead to a substantial distribution shift where local structures are almost entirely lost in large circuits.
Finding a realistic random-circuit generator for large quantum circuits could be interesting for both benchmarking and the training of novel RL-agents.
These are some works in progress and are left to the future.

\section*{Acknowledgments}
We would like to thank K. Staudacher for helpful discussions on the theory of ZX calculus.
The research is part of the Munich Quantum Valley (MQV), which is supported by the Bavarian state government with funds from the Hightech Agenda Bayern Plus.

\appendix

\section{Machine-learning approach}

\subsection{Graph-neural networks}\label{ap:GNN}
Graph Neural Networks (GNN) \cite{Bronstein2021GeometricDL} operate on inputs made up of arbitrarily structured graphs $G=(N,E)$ of nodes $N$ and edges $E$ between nodes.
Most current methods operate on an algorithmic scheme known as `Message Passing' 
where iteratively information in the form of a per-node embedding $e_n$ is transmitted to all of the node's neighbors and combined using a permutation-invariant aggregation function, such as \texttt{sum}, \texttt{mean}, or \texttt{max}, followed by a per-node transformation.
This means that after $K$ steps of message passing, every node is able to observe all other nodes $K$ edges far from itself.
In this work, we specifically use a GCN~\cite{Kipf2016SemiSupervisedCW}, which generalizes the notion of the Convolutional Neural Network~\cite{Lecun1998Convolutional} from image processing to arbitrarily structured neighborhood configurations.
We preprocess our graph by adding `virtual nodes' to every edge which hold information on the ZX-rules applicable to edges.
For features, we give every node - both virtual and real ones - all rules applicable to that node/edge.
This means our approach is naturally invariant to a color change of the entire graph: flipping every node from red to green and vice versa will not change our representation.
This naturally reflects the ZX calculus' property of every rule being available in both colors.

\subsection{Circuit extraction}
\label{Circuit extraction}
Given a ZX diagram with the promise that it represents a unitary map, it is generally hard to extract a circuit if no ancilla qubits are allowed \cite{Beaudrap2022}. Unitarity is a global property of a ZX diagram that can be composed of parts that are not unitary by itself and the non-unitary parts might only cancel globally. The currently most powerful extraction algorithms \cite{Backens2021, Duncan2020} make use of a graph-theoretic property known as \textit{gflow}. If a graph-like diagram (a particular representation of a ZX diagram, introduced below) has this property, extraction can be guaranteed. Since quantum circuits have gflow, powerful simplification algorithms have been developed \cite{Duncan2020, Kissinger2020a, Staudacher2023} based on a restricted set of transformation rules whose application guarantee the preservation of the existence of a gflow.

The circuit extraction algorithm accepts graph-like diagrams as input. A graph-like diagram is a particular representation of a ZX diagram which only consists of Z spiders connected by Hadamard edges
\begin{equation}
\label{Hadamard}
\vcenter{\hbox{\includegraphics[width=2.54cm]{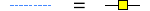}}}\,,
\end{equation}
no self loops and no parallel edges. Any ZX diagram can be converted into a graph-like diagram. This representation allows powerful simplification rules such as \textit{local complementation} to remove spiders with $\pm\pi/2$ phases and \textit{pivoting} \cite{Kotzig1968,Duncan2020} which remove pairs of spiders with phases zero and $\pi$. Pivoting and local complementation additionally modify the adjacency of the local neighborhood of the nodes which can increase the number of connections. These rules (and additional ones) were used in Refs. \cite{Staudacher2023,Holker2024} to decrease the number of connections, guided by rule selection heuristics.
The extraction algorithm \cite{Duncan2020, Backens2021} proceeds from right-to-left through the diagram, extracting single-qubit gates, CZ and CNOT gates whereby the existence of a gflow guarantees success.

The application of a transformation sequence based on the rules of \cref{fig:rules} typically breaks the gflow of a ZX diagram. To still enable circuit extraction, after optimization by the RL agent, we transform the diagram to graph-like form and subsequently apply a pre-processing step before passing the diagram to the extraction method. This procedure is chosen in the way to keep the structure of the extracted circuit as closely as possible to the ZX graph passed to the extraction method. To this end, we propose different \textit{extraction levels}. Our extraction method starts with extraction level 1 which amounts to minimal pre-processing of the ZX graph. The pre-processed graph is subsequently passed to PyZX's extraction method. If extraction fails, we proceed to extraction level 2 etc.  The different extraction levels are defined as:\\
Level 1: The pivoting and local complementation rules are applied to all nodes with exactly two in or outgoing wires. This reverts application of the Euler and $\pi$-commute rule which otherwise might prevent extraction.\\
Level 2-4: The pivoting and local complementation rules are applied to all nodes with 3 (level 2), 4 (level 3) and arbitrary (level 4) in or outgoing wires. This procedure reverts problematic applications of the bialgebra rule.   \\
Level 5: If extraction still fails, PyZX's method full\_reduce is applied.
Although this pre-processing step does not guarantee extraction, we did not observe failure cases during training and evaluation of our algorithm. Moreover, this aspect can be seen as part of the agent's task to pass the ZX diagram to the extraction method in a form that is both extractable and leads to optimized circuits.
Interestingly, transformation of a quantum circuit to a ZX diagram and reconstruction via level 1 extraction already constitutes a powerful CNOT optimizer due to non-trivial CNOT-gate cancellations.

\subsection{Features}\label{sec:Features}
\begin{table}[!h]\label{tab:features}
    \centering
    \begin{tabular}{c}
        \toprule
        \midrule
        gate count\\
        t-gate count\\
        clifford-gate count\\
        two-qubit-gate count\\
        hadamard-gate count\\
        circuit depth\\
        circuit depth cz\\
        number of graph edges\\
        \midrule
        \bottomrule
    \end{tabular}
    \caption{
        List of features used for the MLP extractor used for peephole optimization.
        All features additional get divided by the total number of gates and qubits, 
        to account for the increase in e.g. t-gates expected from larger circuits.
    }
\end{table}
We showcase the features used for the peephole tests in \cref{tab:features}.

\bibliographystyle{IEEEtran}
\bibliography{ZX_GNN_RL}

\end{document}